\documentclass[sigconf]{acmart}

\AtBeginDocument{%
  }

\usepackage{bbding}
\usepackage{cleveref}
\usepackage{multirow}
\usepackage{booktabs}    
\usepackage{multirow}    
\usepackage{graphicx}    
\usepackage{pifont}      
\usepackage{xcolor}      

\usepackage{pifont}
\newcommand{\cmark}{\ding{51}} 
\newcommand{\xmark}{\ding{55}} 

\copyrightyear{2026}
\acmYear{2026}
\setcopyright{cc}
\setcctype{by-nc-nd}
\acmConference[MM '26]{Proceedings of the 34th ACM International Conference on Multimedia}{November 10--14, 2026}{Rio de Janeiro, Brazil}
\acmBooktitle{Proceedings of the 34th ACM International Conference on Multimedia (MM '26), November 10--14, 2026, Rio de Janeiro, Brazil}
\acmDOI{10.1145/3767308.3836543}
\acmISBN{979-8-4007-2213-4/2026/11}
\acmSubmissionID{mfp9431}

\newcommand{\textitgray}[1]{\textcolor{gray}{\textit{#1}}}
\newcommand{\redtext}[1]{\textcolor{red}{#1}}
\newcommand{\greentext}[1]{\textcolor{green!50!black}{#1}}
\newcommand{\first}[1]{\textcolor{red}{\textbf{#1}}}
\newcommand{\second}[1]{\textcolor{blue}{\underline{#1}}}
\newcommand{\third}[1]{\textcolor{orange}{#1}}

\begin{document}

\title{VGGT-Align: Bridging Local Reconstruction and Global Consistency for Long-Sequence 3D Reconstruction}
\author{Wei Zhang}
\email{zhangwei707@mail.nwpu.edu.cn}
\affiliation{%
  \department{School of Computer Science}
  \institution{Northwestern Polytechnical University}
  \city{Xi'an}
  \country{China}}

\author{Yihang Wu}
\email{wuyihang@mail.nwpu.edu.cn}
\affiliation{%
  \institution{Northwestern Polytechnical University}
  \city{Xi'an}
  \country{China}}

\author{Songhua Li}
\email{lisonghua@mail.nwpu.edu.cn}
\affiliation{%
  \institution{Northwestern Polytechnical University}
  \city{Xi'an}
  \country{China}}

\author{Qi Wang}
\email{crabwq@gmail.com}
\affiliation{%
  \institution{Northwestern Polytechnical University}
  \city{Xi'an}
  \country{China}}

\renewcommand{\shortauthors}{Zhang et al.}

\begin{abstract}

  Maintaining global geometric consistency is a central challenge in long-sequence 3D reconstruction, with scale drift being the most critical failure mode. In chunk-based inference pipelines, the scale degree of freedom in sequential Sim(3) alignment is left unconstrained, causing estimation errors to compound multiplicatively and distort global trajectories and point cloud geometry. We present a scale-consistency enhancement framework built on a key insight: in structured environments such as driving scenes, geometric quantities arising from environmental regularity remain inherently invariant across temporal segments, and discrepancies in their per-chunk measurements directly expose inter-chunk scale drift. We propose Scene Geometric Invariant Anchoring (SGIA), which extracts dominant geometric invariants from each chunk's predicted point cloud via coarse-to-fine robust estimation and exploits their cross-chunk consistency to establish scale constraints independent of point cloud registration, explicitly degenerating 7-DoF Sim(3) alignment into 6-DoF rigid-body transformation and severing chain-wise scale error propagation at its source. We further introduce a lightweight test-time adaptation strategy that fine-tunes only normalization-layer parameters via multi-objective self-supervision, progressively improving intra-chunk predictions along the sequence. Both modules are plug-and-play and require no offline retraining. Experiments on multiple long-sequence benchmarks demonstrate state-of-the-art performance, reducing absolute trajectory error by up to 32\% with significant gains in trajectory stability and reconstruction quality. \textbf{Code:}~{\small\url{https://github.com/WZ-CS/VGGT-Align}}.
\end{abstract}


\begin{teaserfigure}
  \centering
  \includegraphics[width=0.9\textwidth]{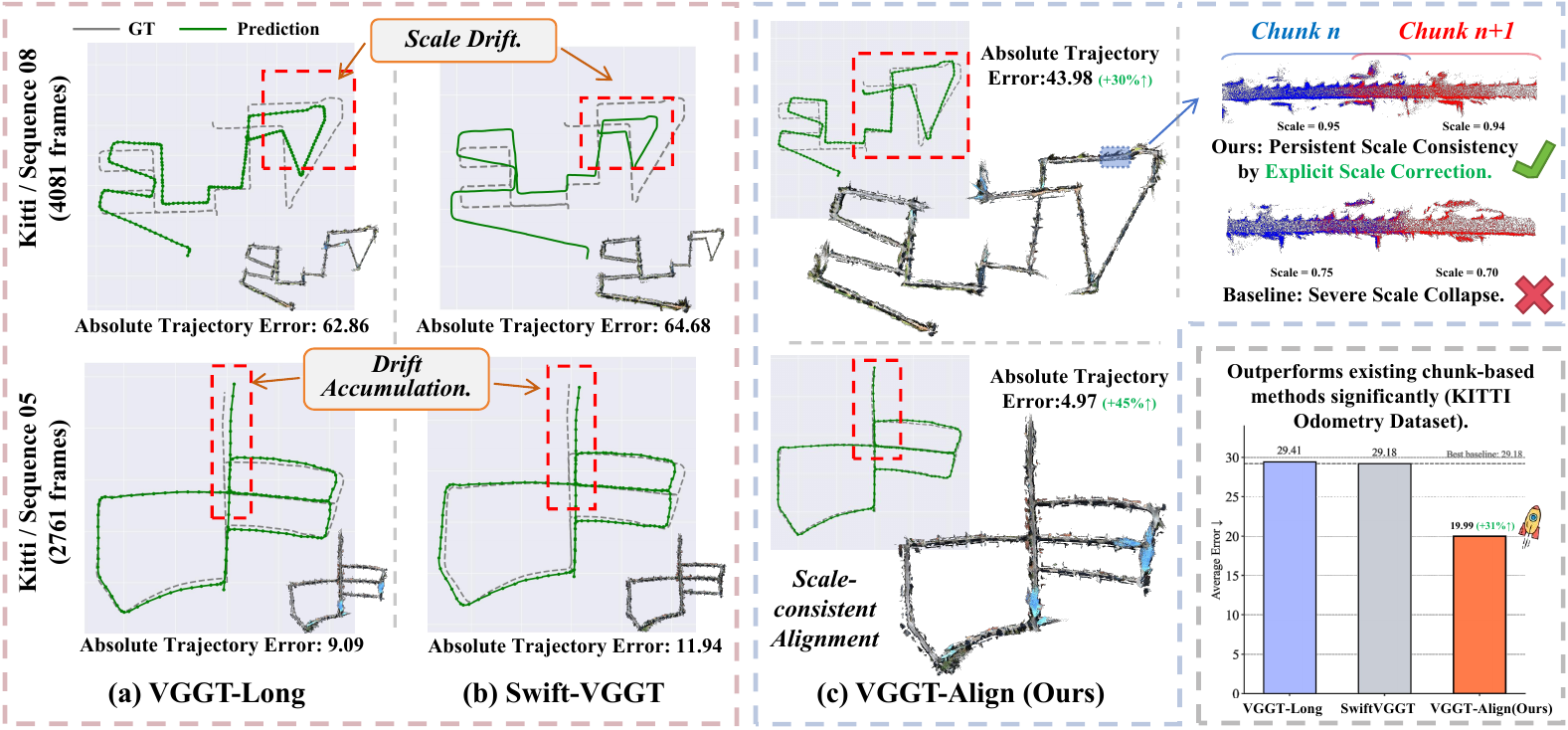}
  \caption{Comparison of chunk-based long-sequence reconstruction on KITTI. Existing baselines (a, b) suffer from unconstrained scale drift in Sim(3) alignment, manifesting as scale drift (trajectory shrinks into a distorted cluster) and drift accumulation (trajectory diverges from ground truth). (c) VGGT-Align maintains scale consistency by anchoring each chunk via scene geometric invariants, reducing ATE by 30\% on Seq 08 and 45\% on Seq 05. Right: inter-chunk point cloud overlay shows our scale correction preserves geometric consistency between adjacent chunk (top, \cmark), while the baseline exhibits severe drift at chunk boundaries (bottom, \xmark). VGGT-Align achieves state-of-the-art among chunk-based methods.}
  \Description{A comparison of long-sequence 3D reconstruction methods on KITTI. Baseline trajectories exhibit accumulated scale drift and geometric misalignment at chunk boundaries, whereas VGGT-Align remains close to the ground-truth trajectory and preserves overlap between adjacent reconstructed point-cloud chunks.}
  \label{fig:teaser}
\end{teaserfigure}

\maketitle

\vspace{-1.6 mm}

\begin{figure}[h]
  \centering
  \includegraphics[width=0.72\linewidth]{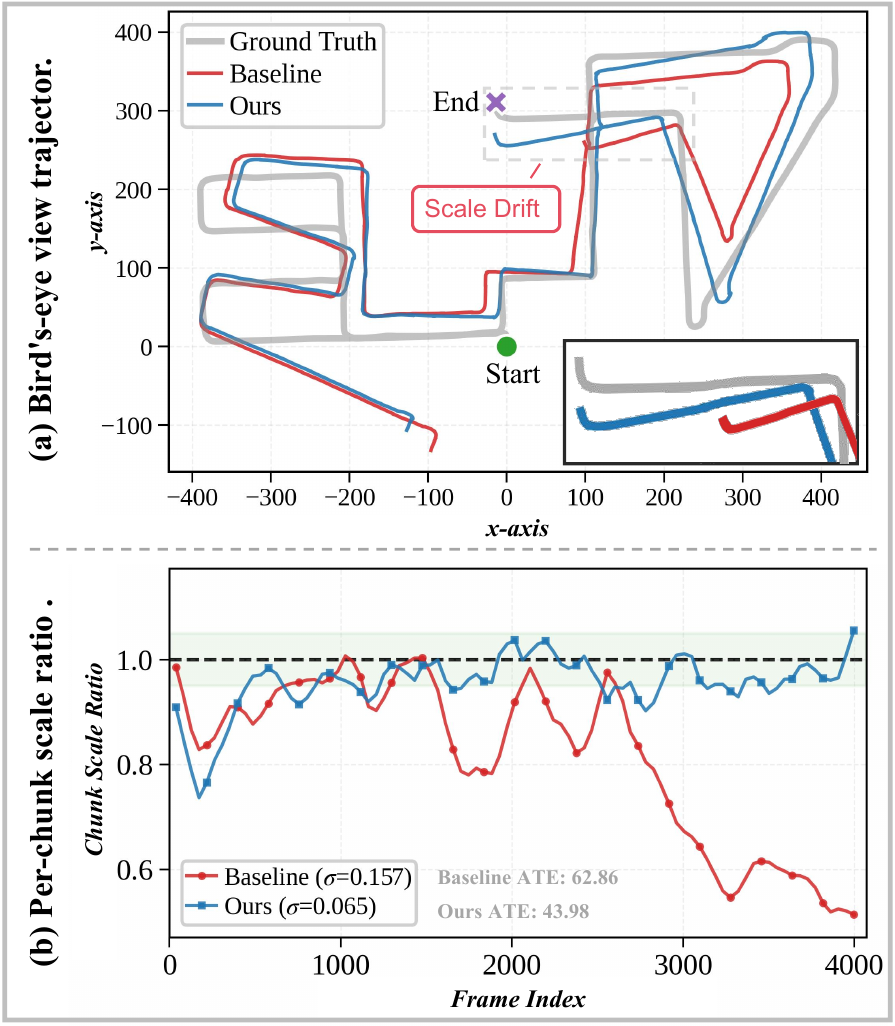}
  \caption{Qualitative analysis of camera pose estimation on KITTI Seq 08. (a) Bird's-eye view trajectory comparison. The baseline exhibits significant scale drift in the later portion of the sequence, causing the estimated trajectory to deviate substantially from the ground truth, while our method maintains closer agreement throughout by introducing global scale prior constraints during cross-chunk alignment. (b) Per-chunk scale consistency after Sim(3) alignment. Each point represents the local path length ratio (predicted / ground truth) within a chunk of 75 frames. $\sigma$ denotes its standard deviation across all chunks. The baseline suffers progressive scale compression, dropping to approximately 0.5 by the end of the sequence ($\sigma$=0.157), while our method maintains the ratio near the ideal value of 1.0 ($\sigma $=0.065).}
\end{figure}

\section{Introduction}
\label{sec:intro}

Feed-forward 3D vision models~\cite{vaswani2017attention, vggsfm, vggt, dust3r, mast3r} have demonstrated capability in recovering dense geometry from unposed images~\cite{10720843,11112633,11068977,11595022,zhang2026vggtcd}. By predicting camera parameters, depth maps, and 3D point clouds in a single forward pass, these methods bypass the iterative optimization of classical SfM and SLAM~\cite{triggs1999bundle, schonberger2016structure}, achieving efficiency on short sequences. However, their fixed context window, typically limited to tens of frames, prevents direct application to long sequences comprising thousands of frames, as encountered in autonomous driving and large-scale mapping.

To bridge this gap, chunk-based methods~\cite{vggt-long, swiftvggt} partition sequences into overlapping segments, process each independently with a feed-forward model, and sequentially align local reconstructions via Sim(3) transformations estimated from overlap regions. While effective locally, this paradigm introduces a critical vulnerability: \textbf{unconstrained scale drift}. The scale factor at each step is estimated solely from noisy overlap consistency, and since each chunk's coordinate frame is defined only relative to its predecessor, scale errors propagate \textit{multiplicatively} along the chain, causing progressive trajectory distortion from local inconsistency to global structural collapse (\cref{fig:teaser}). The root cause is that scale is treated as a free variable without independent constraint. Yet we observe that this is unnecessarily permissive: 
many structured environments, particularly those with rigidly mounted cameras and regular geometry (e.g., driving scenes), exhibit geometric quantities that remain inherently invariant across temporal segments.
Although their per-chunk measurements vary due to unknown local scale, the \textit{ratios} between adjacent chunks directly reveal inter-chunk scale discrepancy, independently of point cloud registration.

Building on this observation, we propose \textbf{VGGT-Align}, a scale-consistency enhancement framework for long-sequence reconstruction. Our primary contribution is \textbf{Scene Geometric Invariant Anchoring (SGIA)}, which extracts dominant geometric structures from each chunk's predicted point cloud via coarse-to-fine robust estimation, and exploits their cross-chunk consistency to establish scale constraints independent of overlap-based alignment. By injecting these constraints, the 7-DoF Sim(3) alignment is explicitly degenerated into a 6-DoF rigid-body transformation, severing chain-wise error propagation at its source. The mechanism operates purely on the model's own predictions, requiring no external sensors and no offline training. As a secondary contribution, we introduce a lightweight \textbf{test-time adaptation (TTA)} strategy that fine-tunes only normalization-layer parameters during inference, using photometric, geometric, and temporal smoothness self-supervision. This enables the model to progressively adapt to the current scene, improving intra-chunk prediction quality and complementing inter-chunk scale anchoring. Our contributions are summarized as follows:
\begin{itemize}
    \item We identify and formally characterize scale drift in chunk-based reconstruction, showing it arises from unconstrained scale in sequential Sim(3) alignment and compounds multiplicatively.
    \item We propose SGIA, which leverages inherent scene structural regularity to provide per-chunk scale constraints, degenerating Sim(3) into rigid-body transformation and eliminating scale drift at its source.
    \item We introduce a complementary test-time adaptation strategy that improves intra-chunk predictions via online normalization layer fine-tuning.
    \item Extensive experiments on long-sequence benchmarks demonstrate that VGGT-Align achieves state-of-the-art performance, reducing average trajectory error by up to 30\% over existing methods, with consistent improvements in trajectory stability, scale consistency, and reconstruction quality.
\end{itemize}

\section{Related Work}
\label{sec:related}
\noindent\textbf{Feed-forward 3D models and long-sequence reconstruction.}
Feed-forward transformers such as DUSt3R~\cite{dust3r}, MASt3R~\cite{mast3r}, and VGGT~\cite{vggt} predict dense 3D quantities in a single pass, but are limited by their fixed context window and run out of memory on long sequences. Fast3R~\cite{fast3r} extends to larger frame sets but still cannot handle thousands of frames on consumer hardware. To address this, chunk-based methods partition videos into overlapping segments and stitch local reconstructions via Sim(3) alignment. VGGT-Long~\cite{vggt-long} introduced this paradigm with IRLS alignment and loop closure; SwiftVGGT~\cite{swiftvggt} accelerated it via reliability-guided sampling and internal token reuse for loop detection~\cite{nister2004efficient, rublee2011orb, sarlin2020superglue, lindenberger2023lightglue}. Other scaling strategies include real-time SLAM backends~\cite{mast3r-slam,vggt-slam, vggt-motion}, persistent 3D states~\cite{cut3r}, causal KV caching~\cite{streamvggt}, and dynamic scene extensions and robust reconstruction benchmarks~\cite{vggt4d,skyevents}. However, all sequential alignment approaches leave the scale degree of freedom unconstrained, leading to multiplicative drift that our work directly addresses.
\vspace{2pt}

\noindent\textbf{Scale priors and test-time adaptation.}
Recovering metric scale from monocular has been a longstanding challenge, approached through ground plane constraints~\cite{zhou2019ground}, known object dimensions and structural priors~\cite{song2015joint, yang2019objectplane}, IMU fusion~\cite{campos2021orb}, and metric depth networks~\cite{bhat2023zoedepth, yin2023metric3d, hu2024metric3dv2}. Self-supervised monocular depth methods~\cite{godard2019digging, guizilini2023zerodepth} also exploit vehicle velocity or stereo baselines as scale signals during training. Our method differs by operating on predicted 3D point clouds at inference time and requiring only cross-chunk \textit{consistency} of geometric invariants rather than known absolute values. For test-time adaptation~\cite{zhao2026sharp,10847788,JIA2025265,11184567}, prior works address domain shift in depth estimation~\cite{tonioni2019unsupervised,kuznietsov2021comoda}, stereo~\cite{tonioni2019real}, and image classification~\cite{wang2020tent,xiao2026staying,xiao2026layer,xiao2026not,xiao2025visual}. We adopt a similar strategy tailored to chunk-based reconstruction, carrying adapted parameters forward across chunks to progressively improve predictions~\cite{lyu2025towards,m2d_lgt_dance_tempomoe,lyucome}.

\begin{figure*}[t]
\centering
\includegraphics[width=0.8\linewidth]{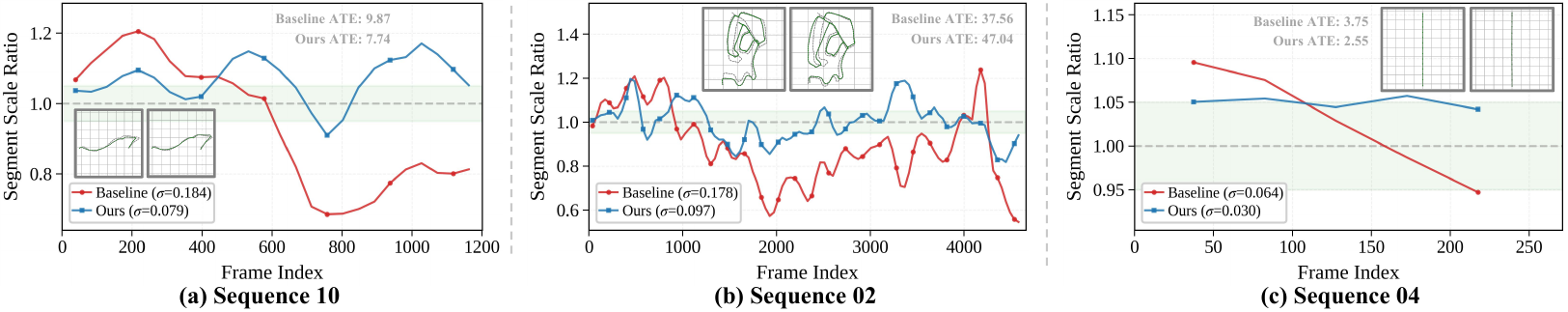} 
\caption{Per-chunk scale ratio analysis across three KITTI sequences of varying length. The ideal ratio is 1.0 (dashed line); the green band denotes $\pm$5\% tolerance. The baseline (red) exhibits systematic scale drift with high variance, while VGGT-Align (blue) maintains ratios tightly around 1.0 with consistently lower variance (e.g., $\sigma$: 0.184$\to$0.079 on Seq~10, 0.064$\to$0.030 on Seq~04). Inset trajectory plots show corresponding ATE improvements. On Seq~02, scale correction is visibly effective ($\sigma$: 0.178$\to$0.097), though the trajectory exhibits a slight positional offset due to residual rotational drift in this 5\,km loop; see supplementary material for extended visualizations.}
\label{fig:fig_vis_1}
\end{figure*}

\section{Method}
\label{sec:method}

\subsection{Preliminaries and Problem Formulation}
\label{sec:prelim}

\noindent\textbf{Chunk-based long-sequence reconstruction.}
Given a long image sequence $\mathcal{I} = \{I_1, \ldots, I_N\}$, we partition it into $M$ overlapping chunks $\{C_k\}_{k=1}^{M}$ with window size $L$ and overlap $O$. Each chunk is processed by a feed-forward model $\mathcal{F}$ (e.g., VGGT~\cite{vggt}), outputting a local point cloud $\mathcal{P}_k \in \mathbb{R}^{L \times H \times W \times 3}$ with per-pixel confidence $\mathcal{W}_k$, camera extrinsics $\mathbf{T}_k^{(i)}$, and intrinsics $\mathbf{K}_k^{(i)}$. Adjacent chunks are stitched by estimating a similarity transformation $(s_k, \mathbf{R}_k, \mathbf{t}_k) \in \mathrm{Sim}(3)$ from the overlap region via confidence-weighted IRLS~\cite{vggt-long}:
\begin{equation}
\label{eq:irls}
    s_k^*, \mathbf{R}_k^*, \mathbf{t}_k^* = \underset{s, \mathbf{R}, \mathbf{t}}{\arg\min} \sum_{i \in \mathcal{O}} w_i \, \rho \big( \| \mathbf{p}_i^{(k)} - (s \mathbf{R} \mathbf{p}_i^{(k+1)} + \mathbf{t}) \| \big),
\end{equation}
where $\mathcal{O}$ indexes overlapping frames, $w_i$ is point confidence, and $\rho(\cdot)$ is a robust loss.

\vspace{2pt}
\noindent\textbf{The scale drift problem.}
Transforming chunk $C_k$ into the global frame of $C_1$ requires composing all intermediate transforms, yielding a cumulative scale $\bar{s}_k = \prod_{j=1}^{k-1} s_j$. Since each $s_j = s_j^*(1 + \epsilon_j)$ carries estimation error $\epsilon_j$, the cumulative scale becomes:
\begin{equation}
\label{eq:drift}
    \bar{s}_k = \bar{s}_k^* \prod_{j=1}^{k-1} (1 + \epsilon_j).
\end{equation}
Even a modest per-step bias ($\mathbb{E}[\epsilon_j] = 0.02$) compounds exponentially: $1.02^{50} \approx 2.7\times$ after 50 chunks. We verify this empirically in \cref{fig:fig_vis_1}: the baseline's per-chunk scale ratio deviates systematically from 1.0 with high variance ($\sigma = 0.184$), confirming compounding bias rather than zero-mean noise. This motivates our core contribution: an independent scale constraint decoupled from overlap-based registration.

\subsection{Framework Overview}
\label{sec:overview}
 
\begin{figure*}[t!]
\centering
\includegraphics[width=0.85\linewidth]{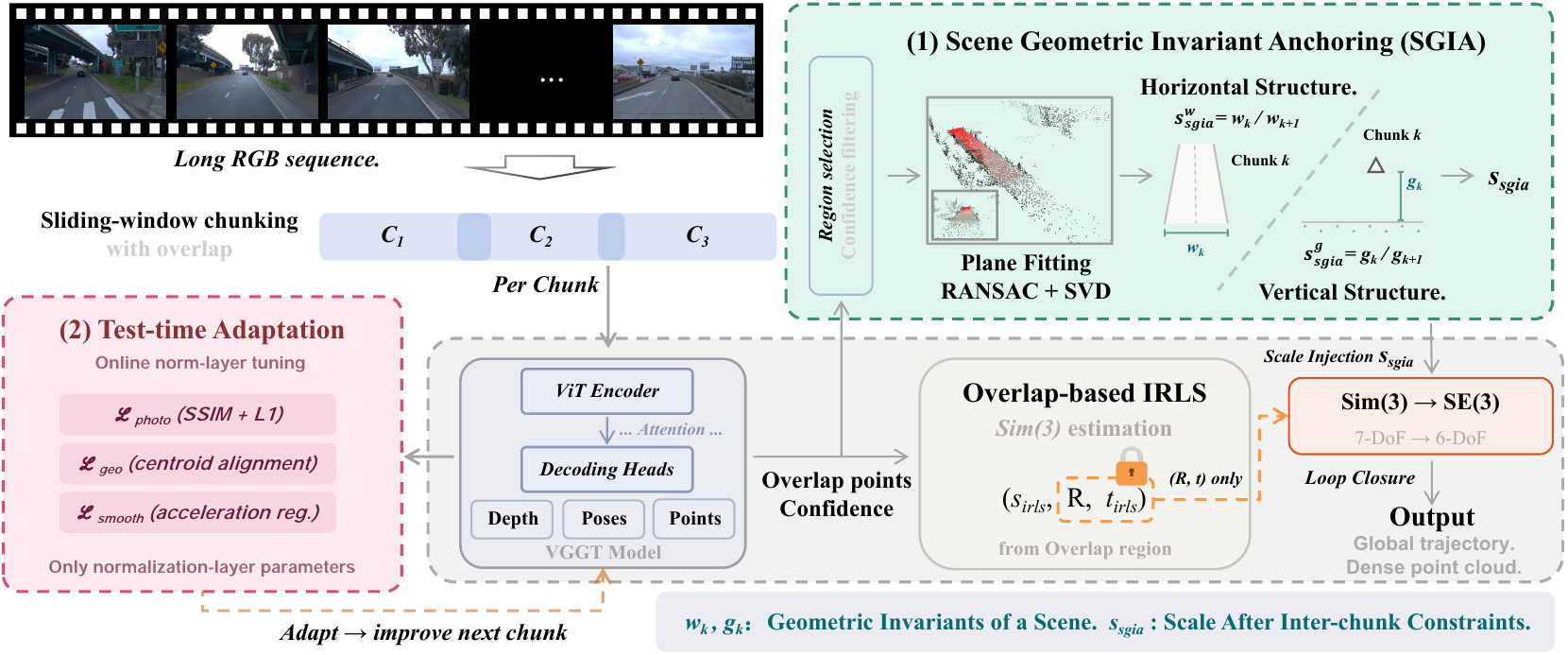} 
\caption{Overview of the VGGT-Align framework. A long RGB sequence is partitioned into overlapping chunks and processed independently by a feed-forward 3D model. Our two plug-and-play contributions are highlighted: \textbf{(1)} Scene Geometric Invariant Anchoring (SGIA, green, top-right) extracts dominant geometric structures from each chunk's predicted point cloud via region selection with confidence filtering, followed by coarse-to-fine plane estimation (RANSAC + SVD). Two complementary invariant types are exploited: vertical structures (e.g., camera-to-ground distance $g_k$) and horizontal structures (e.g., road width $w_k$), whose cross-chunk measurement ratios yield per-pair scale constraints $s_{\text{sgia}}$ independent of point cloud registration. This constraint is injected into the overlap-based IRLS alignment, explicitly replacing the estimated scale factor (lock symbol) and degenerating the 7-DoF Sim(3) transformation into a 6-DoF rigid-body transform SE(3), thereby severing chain-wise scale drift at its source. \textbf{(2)} Test-time adaptation (TTA, pink, bottom-left) fine-tunes only normalization-layer parameters using three self-supervised objectives: photometric consistency, geometric alignment, and temporal smoothness, with the adapted parameters improving predictions for subsequent chunks (dashed feedback arrow). The aligned chunks are further refined by loop closure optimization to produce the final globally consistent trajectory and dense point cloud.}
\label{fig:overview}
\end{figure*}
 
An overview of VGGT-Align is shown in \cref{fig:overview}. The pipeline proceeds as follows: \textbf{(1)}~The input sequence is partitioned into overlapping chunks via a sliding window. \textbf{(2)}~Each chunk is processed by a feed-forward 3D model to obtain per-chunk predictions (depth, poses, and 3D points). Prior to inference, TTA optionally adapts the model's normalization-layer parameters using self-supervised objectives from the previous chunk, progressively improving prediction quality along the sequence. \textbf{(3)}~For each chunk, SGIA extracts geometric invariant observations from the predicted point cloud. \textbf{(4)}~During pairwise alignment, the SGIA-derived scale constraint is injected into the IRLS result, explicitly replacing the estimated scale factor and degenerating the $\mathrm{Sim}(3)$ alignment into a rigid-body $\mathrm{SE}(3)$ transformation. \textbf{(5)}~An optional loop closure module further refines global consistency. The two proposed modules are complementary: SGIA operates at the \textit{inter-chunk alignment} stage, constraining the scale degree of freedom; TTA operates at the \textit{intra-chunk inference} stage, improving prediction quality. Both are plug-and-play and require no offline retraining.

\subsection{Scene Geometric Invariant Anchoring}
\label{sec:sgia}

\subsubsection{Geometric Invariant Observation}
\label{sec:invariant_obs}

The key insight behind SGIA is the following: natural scenes contain geometric quantities that are \textit{inherently constant} in the physical world but whose measured values in each chunk's local coordinate system vary due to the unknown, chunk-specific scale factor. The discrepancy between measurements from adjacent chunks reveals their relative scale offset, \textit{independently of} the overlap-based point cloud registration.

Formally, let $g \in \mathbb{R}^+$ be a geometric quantity that is constant across all chunks. In chunk $C_k$'s local coordinate system, this quantity is measured as $g_k$, related to the true value through the chunk's implicit absolute scale $\sigma_k$:
\begin{equation}
\label{eq:invariant_def}
    g_k = g \,/\, \sigma_k.
\end{equation}
For adjacent chunks $C_k$ and $C_{k+1}$, the ratio of their measurements gives the true relative scale:
\begin{equation}
\label{eq:relative_scale}
    \frac{g_k}{g_{k+1}} = \frac{\sigma_{k+1}}{\sigma_k} = s_{k \to k+1}^*.
\end{equation}
Crucially, \cref{eq:relative_scale} holds for \textit{any} geometric invariant $g$ and does \textit{not} require knowing the absolute value of $g$, only that it remains constant across chunks. This makes the constraint broadly applicable without external calibration.
 
We identify two complementary classes of geometric invariants prevalent in structured environments:
 
\vspace{2pt}
\noindent\textbf{Vertical invariants} capture the relationship between the camera and a dominant horizontal plane. The most natural instance is the camera-to-ground distance $g_k$, which remains constant when the camera is rigidly mounted, as in vehicle-mounted settings.
 
\vspace{2pt}
\noindent\textbf{Horizontal invariants} capture lateral scene structure that persists across chunks. In driving scenarios, road width $w_k$, determined by lane markings and road boundaries, is locally constant over extended stretches.
 
Both types provide independent scale observations per chunk. When multiple sources are available, they can be fused via weighted combination to improve robustness (\cref{sec:scale_inject}).

\subsubsection{Robust Invariant Extraction}
\label{sec:extraction}
 
The extraction pipeline consists of three stages: candidate region selection, coarse-to-fine plane estimation, and invariant measurement.
 
\vspace{2pt}
\noindent\textbf{Stage 1: Candidate region selection.}
For vertical invariants, we exploit the spatial prior that ground surfaces project onto the lower portion of the image. For each frame in chunk $C_k$, we select 3D points corresponding to the bottom $\rho$ fraction of image rows (default $\rho = 0.3$) as ground candidates, and further filter by retaining only points whose confidence exceeds the $\tau$-th percentile (default $\tau = 40$). For horizontal invariants, a similar strategy selects points from a central horizontal band.
 
\vspace{2pt}
\noindent\textbf{Stage 2: Coarse-to-fine plane estimation.}
We fit a plane $\mathbf{n}^\top \mathbf{x} + d = 0$ to the aggregated candidates via a two-stage procedure. In the coarse stage, RANSAC~\cite{fischler1981random} identifies an initial inlier set. In the refinement stage, we compute the covariance matrix of the inliers and extract the plane normal as the eigenvector of the smallest eigenvalue via SVD:
\begin{equation}
\label{eq:svd_plane}
    \mathbf{C} = (\mathbf{X}_{\text{in}} - \bar{\mathbf{x}})^\top (\mathbf{X}_{\text{in}} - \bar{\mathbf{x}}), \quad
    \mathbf{n}^* = \mathbf{v}_{\min}(\mathbf{C}),
\end{equation}
where $\mathbf{X}_{\text{in}}$ denotes the inlier point matrix, $\bar{\mathbf{x}}$ is the inlier centroid, and $\mathbf{v}_{\min}(\cdot)$ returns the eigenvector of the smallest eigenvalue. This combines the outlier robustness of RANSAC with the statistical optimality of PCA-based fitting.
 
\vspace{2pt}
\noindent\textbf{Stage 3: Invariant measurement.}
Given the estimated plane $(\mathbf{n}^*, d^*)$, we compute the signed distance from each frame's camera center $\mathbf{c}_f = \mathbf{T}_k^{(f)}[:3, 3]$ to the plane:
\begin{equation}
\label{eq:height}
    h_f = \mathbf{n}^{*\top} \mathbf{c}_f + d^*.
\end{equation}
The vertical invariant for chunk $C_k$ is the median over valid frames:
\begin{equation}
\label{eq:gk}
    g_k = \mathrm{median}\{h_f \mid h_f > 0, \; f = 1, \ldots, |C_k|\}.
\end{equation}
The median provides robustness to outlier frames caused by partial occlusion or non-planar regions. For horizontal invariants, $w_k$ is computed analogously as the median lateral extent of ground-plane inliers perpendicular to the estimated forward motion.

\subsubsection{Scale Injection and Alignment Degeneracy}
\label{sec:scale_inject}
 
Given invariant observations from adjacent chunks, we compute the prior-based relative scale. For a single invariant source:
\begin{equation}
\label{eq:s_sgia_single}
    s_{\text{sgia}} = g_k \,/\, g_{k+1}.
\end{equation}
When multiple sources are available (e.g., both vertical $g_k$ and horizontal $w_k$), we fuse them via weighted combination:
\begin{equation}
\label{eq:s_sgia_multi}
    s_{\text{sgia}} = \lambda_g \cdot \frac{g_k}{g_{k+1}} + \lambda_w \cdot \frac{w_k}{w_{k+1}}, \quad \lambda_g + \lambda_w = 1,
\end{equation}
where the weights reflect relative reliability (default $\lambda_g = 0.7$, $\lambda_w = 0.3$).
 
\vspace{2pt}
\noindent\textbf{Scale replacement.}
The IRLS alignment (\cref{eq:irls}) yields $(s_{\text{irls}}, \mathbf{R}, \mathbf{t}_{\text{irls}})$. We replace the estimated scale with the prior-derived value:
\begin{equation}
\label{eq:blend}
    s = \alpha \cdot s_{\text{sgia}} + (1 - \alpha) \cdot s_{\text{irls}},
\end{equation}
where $\alpha \in [0, 1]$ controls the prior strength. We find $\alpha = 1$ optimal on KITTI, while $\alpha \in [0.7, 0.9]$ works better on datasets with less regular geometry.
 
Since the rotation $\mathbf{R}$ is retained from IRLS, the translation must be adjusted to preserve centroid alignment of the overlap region:
\begin{equation}
\label{eq:t_adjust}
    \mathbf{t} = \mathbf{t}_{\text{irls}} + (s_{\text{irls}} - s) \cdot \mathbf{R} \bar{\mathbf{p}}_2,
\end{equation}
where $\bar{\mathbf{p}}_2$ is the overlap centroid in chunk $C_{k+1}$'s frame.
 
\vspace{2pt}
\noindent\textbf{Alignment degeneracy: $\mathrm{Sim}(3) \to \mathrm{SE}(3)$.}
When $\alpha = 1$, the scale is fully determined by the geometric invariant, independent of overlap-based registration. The 7-DoF $\mathrm{Sim}(3)$ alignment degenerates into 6-DoF $\mathrm{SE}(3)$. This breaks the multiplicative error chain in \cref{eq:drift}: the cumulative scale is now determined by per-chunk invariant ratios rather than compounding noisy IRLS estimates. Since each chunk's invariant is measured independently, errors do not accumulate.
 
\vspace{2pt}
\noindent\textbf{Robustness and fallback.}
When invariant extraction fails for a particular chunk (e.g., insufficient ground points), the system falls back to the IRLS-only scale by setting $\alpha = 0$. An adaptive blending strategy can further adjust $\alpha$ based on extraction confidence (RANSAC inlier count, per-frame height variance), providing graceful degradation.

\subsection{Test-Time Adaptation}
\label{sec:tta}
 
While SGIA addresses inter-chunk alignment, per-chunk prediction quality can degrade over long sequences due to distributional shift between training data and the test scene. We introduce a lightweight test-time adaptation (TTA) mechanism that enables the model to progressively adapt to the current scene.
 
\vspace{2pt}
\noindent\textbf{Parameter selection.}
To avoid catastrophic forgetting, we restrict adaptation to \textit{normalization-layer parameters only} (LayerNorm scale and bias). These constitute a negligible fraction of total parameters but modulate feature distributions at every layer, making them effective targets for distribution alignment. All other parameters remain frozen.
 
\vspace{2pt}
\noindent\textbf{Self-supervised objectives.}
We employ three complementary losses computed without ground-truth supervision:
 
\vspace{1pt}
\noindent (1) \textit{Photometric consistency} $\mathcal{L}_{\text{photo}}$: we warp source frames to target views using predicted depth and poses, and measure reconstruction error with a combination of SSIM and L1:
\begin{equation}
    \mathcal{L}_{\text{photo}} = \frac{1}{|\mathcal{N}|} \sum_{(i,j) \in \mathcal{N}} \big( 0.85 \cdot \mathcal{L}_{\text{SSIM}}(I_i^w, I_j) + 0.15 \cdot \|I_i^w - I_j\|_1 \big),
\end{equation}
where $\mathcal{N}$ is the set of adjacent frame pairs and $I_i^w$ denotes frame $i$ warped to view $j$.
 
\vspace{1pt}
\noindent (2) \textit{Geometric consistency} $\mathcal{L}_{\text{geo}}$: we penalize the confidence-weighted centroid discrepancy between point clouds from neighboring frames.
 
\vspace{1pt}
\noindent (3) \textit{Temporal smoothness} $\mathcal{L}_{\text{smooth}}$: we regularize the second-order derivative of predicted camera translations:
\begin{equation}
    \mathcal{L}_{\text{smooth}} = \frac{1}{N\!-\!2} \sum_{i=1}^{N-2} \| \mathbf{t}_{i+2} - 2\mathbf{t}_{i+1} + \mathbf{t}_i \|^2.
\end{equation}
 
The total loss is $\mathcal{L}_{\text{tta}} = \lambda_p \mathcal{L}_{\text{photo}} + \lambda_g \mathcal{L}_{\text{geo}} + \lambda_s \mathcal{L}_{\text{smooth}}$, with default weights $\lambda_p\!=\!1.0$, $\lambda_g\!=\!0.5$, $\lambda_s\!=\!0.1$.
 
\vspace{2pt}
\noindent\textbf{Adaptation protocol.}
For each chunk $C_k$ (after an initial warmup), we perform $T$ gradient steps (default $T\!=\!3$) on $\mathcal{L}_{\text{tta}}$ with learning rate $10^{-4}$, Adam optimization~\cite{kingma2014adam}, and gradient clipping. To reduce memory overhead, TTA operates on a subset of $n$ frames (default $n\!=\!3$) at reduced resolution (336px vs.\ 518px for inference), consuming approximately 70\% less VRAM.
 
The adaptation does not alter the current chunk's predictions; the updated parameters take effect during the \textit{next} chunk's forward pass. Parameters are accumulated across chunks by default, enabling progressive specialization to the scene. This adapt-then-infer protocol ensures each chunk benefits from self-supervised signals of all preceding chunks without re-inference.

\begin{table*}[t]
  \centering
 \resizebox{0.95\linewidth}{!}{%
    \begin{tabular}{l|ccc|c|ccccccccccc}
    \toprule
        \textbf{Methods} & \textbf{LC} & \textbf{Calibration} & \textbf{Recon.} & \textbf{Avg.} & \textbf{00} & \textbf{01} & \textbf{02} & \textbf{03} & \textbf{04} & \textbf{05} & \textbf{06} & \textbf{07} & \textbf{08} & \textbf{09} & \textbf{10}   \\ \midrule

        \textitgray{seq. frames} & \textitgray{-} & \textitgray{-}&\textitgray{-}&	\textitgray{2109} &	\textitgray{4542}&	\textitgray{1101}&	\textitgray{4661}&	\textitgray{801}&	\textitgray{271}&	\textitgray{2761}&	\textitgray{1101}&	\textitgray{1101}&	\textitgray{4071}&	\textitgray{1591}&	\textitgray{1201}\\
        
        \textitgray{seq. length (m)} & \textitgray{-} & \textitgray{-}& \textitgray{-} & \textitgray{2012} & \textitgray{3724} & \textitgray{2453} & \textitgray{5067} & \textitgray{561} & \textitgray{394} & \textitgray{2206} & \textitgray{1233} & \textitgray{650} & \textitgray{3223} & \textitgray{1705} & \textitgray{920} \\

        \textitgray{contains loop} & \textitgray{-} & \textitgray{-} & \textitgray{-}& \textitgray{-} & \textitgray{\ding{51}} & \textitgray{\ding{55}} & \textitgray{\ding{51}} & \textitgray{\ding{55}} & \textitgray{\ding{55}} & \textitgray{\ding{51}} & \textitgray{\ding{51}} & \textitgray{\ding{51}} & \textitgray{\ding{55}} & \textitgray{\ding{51}} & \textitgray{\ding{55}} \\
        
        \midrule
        
        DROID-VO \cite{droidslam} & \redtext{\ding{55}} & \redtext{\textit{Required}}& \greentext{\textit{Dense}}  & 54.19 & 98.43  & 84.20  & 108.80  & 2.58  & 0.93  & 59.27  & 64.40  & 24.20  & \third{64.55}  & 71.80  & 16.91   \\ 
        
        DPVO \cite{dpvo}& \redtext{\ding{55}} & \redtext{\textit{Required}}& \redtext{\textit{Sparse}}  & 53.61 & 113.21  & \third{12.69}  & 123.40  & \first{2.09}  & \first{0.68}  & 58.96  & 54.78  & 19.26  & 115.90  & 75.10  & \second{13.63}     \\ 
        
        DROID-SLAM \cite{droidslam} & - & \redtext{\textit{Required}}& \greentext{\textit{Dense}}  & 100.28 & 92.10  & 344.60  & 107.61 & \second{2.38}  & 1.00  & 118.50  & 62.47  & 21.78  & 161.60  & 72.32 & 118.70   \\ 
        
        DPV-SLAM \cite{dpv-slam} & \greentext{\ding{51}} & \redtext{\textit{Required}}& \redtext{\textit{Sparse}}  & 53.03 & 112.80  & \first{11.50}  & 123.53  & \third{2.50}  & \third{0.81}  & 57.80  & 54.86  & 18.77  & 110.49  & 76.66  & \third{13.65}   \\
        
        DPV-SLAM++ \cite{dpv-slam} & \greentext{\ding{51}} & \redtext{\textit{Required}}& \redtext{\textit{Sparse}}  & \second{25.75} & \third{8.30}  & \second{11.86}  & \third{39.64}  & \third{2.50}  & \second{0.78}  & \second{5.74}  & 11.60  & \first{1.52}  & 110.90  & 76.70  & 13.70    \\ 

        \cmidrule(lr){1-16}
        
        MASt3R-SLAM \cite{mast3r-slam} & \greentext{\ding{51}} & \greentext{\textit{No Need}}& \greentext{\textit{Dense}} &	/ & \textitgray{TL} &	\textitgray{TL} & \textitgray{TL} & \textitgray{TL} & \textitgray{TL} & \textitgray{TL} & \textitgray{TL}  & \textitgray{TL} & \textitgray{TL} & \textitgray{TL} & \textitgray{TL}  \\ 
        CUT3R \cite{cut3r} & \redtext{\ding{55}} & \greentext{\textit{No Need}}& \greentext{\textit{Dense}} &	/ & \textitgray{OOM} &	\textitgray{OOM} & \textitgray{OOM} & 148.07 & 22.31 & \textitgray{OOM} & \textitgray{OOM}  & \textitgray{OOM} & \textitgray{OOM} & \textitgray{OOM} & \textitgray{OOM}  \\ 
        Fast3R \cite{fast3r} & \redtext{\ding{55}} & \greentext{\textit{No Need}}& \greentext{\textit{Dense}} &	/ & \textitgray{OOM} &	\textitgray{OOM} & \textitgray{OOM} & \textitgray{OOM} & \textitgray{OOM} & \textitgray{OOM} & \textitgray{OOM}  & \textitgray{OOM} & \textitgray{OOM} & \textitgray{OOM} & \textitgray{OOM}  \\ 
        
        VGGT \cite{vggt} & \redtext{\ding{55}} & \greentext{\textit{No Need}}& \greentext{\textit{Dense}} &	/ & \textitgray{OOM} &	\textitgray{OOM} & \textitgray{OOM} & \textitgray{OOM} & \textitgray{OOM} & \textitgray{OOM} & \textitgray{OOM}  & \textitgray{OOM} & \textitgray{OOM} & \textitgray{OOM} & \textitgray{OOM}  \\ 

        \cmidrule(lr){1-16}
        
        VGGT-Long \cite{vggt-long} & \greentext{\ding{51}} & \greentext{\textit{No Need}}& \greentext{\textit{Dense}} &	29.41 &   9.87& 	111.06& 	\second{37.56}& 	4.89& 	3.75& 	\third{9.09}& 	\second{7.47}& 	\second{4.02}& 	\second{62.86}& 	\third{47.48}& 	25.49   \\ 
        SwiftVGGT \cite{swiftvggt} & \greentext{\ding{51}} & \greentext{\textit{No Need}}& \greentext{\textit{Dense}} &	\third{29.18} &   \second{8.17}& 	102.53& 	\first{36.49}& 	8.12& 	4.88& 	11.94& 	\third{8.88}& 	5.01& 	64.68& 	\second{44.13}& 	26.18   \\ 
        \textbf{VGGT-Align (Ours)} & \greentext{\ding{51}} & \greentext{\textit{No Need}}& \greentext{\textit{Dense}} &	\first{19.99} &   \first{7.74}& 	65.11& 	47.04& 	5.12& 	2.55& 	\first{4.97}& 	\first{4.21}& 	\third{4.48}& 	\first{43.98}& 	\first{23.11}& 	\first{11.66}   \\ 
        \bottomrule
        
    \end{tabular}
  }
    \caption{Camera tracking results (ATE RMSE [m] $\downarrow$) on the KITTI Odometry benchmark. \textitgray{LC}: loop closure capability. VGGT-Align achieves the best overall accuracy among all calibration-free or calibration-required methods. \textitgray{OOM}: CUDA Out-Of-Memory on a single RTX 4090. \textitgray{TL}: Tracking Lost. Color: \first{1st}, \second{2nd}, \third{3rd}.}
  \label{table:kitti_ate}
\end{table*}

\begin{table*}[t]
  \centering
 \resizebox{0.95\linewidth}{!}{%
\begin{tabular}{l|c|c|ccccccccc}
\toprule
\textbf{Methods}       & \textbf{Calib.}   & \textbf{Avg.} & \textbf{163453191} & \textbf{183829460} & \textbf{315615587} & \textbf{346181117} & \textbf{371159869} & \textbf{405841035} & \textbf{460417311} & \textbf{520018670} & \textbf{610454533} \\
\midrule
\textitgray{Frames / Length (m)}   & \textitgray{-}        & \textitgray{198 / 173}      & \textitgray{198 / 160}            & \textitgray{199 / 42}             & \textitgray{199 / 165}            & \textitgray{199 / 351}            & \textitgray{196 / 273}            & \textitgray{199 / 86}             & \textitgray{198 / 266}            & \textitgray{199 / 135}            & \textitgray{198 / 63}             \\
\midrule
DROID-SLAM \cite{droidslam}      & \redtext{\textit{Required}} & 4.396        & 3.705              & \first{0.301}              & \first{0.447}              & 8.653              & 9.320              & 7.621              & 4.170              & \textitgray{TL}                 & \first{0.264}              \\
MASt3R-SLAM \cite{mast3r-slam}     & \greentext{\textit{No Need}}  & 5.560        & 4.500              & \second{0.556}              & \third{1.833}              & 12.544             & 8.601              & \second{1.412}              & 5.428              & 7.910              & \second{1.195}              \\
CUT3R \cite{cut3r}           & \greentext{\textit{No Need}}  & 9.872        & 8.781              & 3.810              & 5.790              & 24.015             & 13.070             & 7.261              & 13.206             & 8.597              & 3.229              \\
Fast3R \cite{fast3r}           & \greentext{\textit{No Need}}  & /            & \textitgray{OOM}                & \textitgray{OOM}                & \textitgray{OOM}                & \textitgray{OOM}                & \textitgray{OOM}                & \textitgray{OOM}                & \textitgray{OOM}                & \textitgray{OOM}                & \textitgray{OOM}                \\
VGGT \cite{vggt}            & \greentext{\textit{No Need}}  & /            & \textitgray{OOM}                & \textitgray{OOM}                & \textitgray{OOM}                & \textitgray{OOM}                & \textitgray{OOM}                & \textitgray{OOM}                & \textitgray{OOM}                & \textitgray{OOM}                & \textitgray{OOM}                \\
\midrule
VGGT-Long \cite{vggt-long} & \greentext{\textit{No Need}}  & \third{3.085}        & \second{3.086}              & 2.544              & 2.336              & \third{4.007}              & \second{4.045}              & 3.141              & \third{2.867}              & \third{3.407}              & 2.333      \\
SwiftVGGT \cite{swiftvggt} & \greentext{\textit{No Need}}  & \second{2.854}        & \third{3.106}              & \third{2.447}              & 2.346              & \first{2.719}              & \third{4.080}              & \third{3.106}              & \second{2.841}              & \second{2.832}              & 2.210      \\
\textbf{VGGT-Align (Ours)} & \greentext{\textit{No Need}}  & \first{1.849}        & \first{1.160}              & 2.540              & \second{0.537}              & \second{3.140}              & \first{2.850}              & \first{1.020}              & \first{1.520}              & \first{2.170}              & \third{1.708}      \\
\bottomrule
\end{tabular}
  }
  \caption{Camera tracking results (ATE RMSE [m] $\downarrow$) on the Waymo Open Dataset. VGGT-Align achieves the best average accuracy among all calibration-free methods and surpasses calibration-required DROID-SLAM. Color: \first{1st}, \second{2nd}, \third{3rd}.}
  \label{table:waymo_ate}
\end{table*}

\section{Experiments}
\label{sec:exp}

\subsection{Experimental Setup}
\label{sec:setup}
 
We evaluate on three driving benchmarks: KITTI Odometry~\cite{geiger2012kitti}, Waymo Open Dataset~\cite{sun2020waymo}, and Virtual KITTI~\cite{gaidon2016vkitti}, comparing against both calibration-required and calibration-free baselines. For tracking we report ATE RMSE (m); for reconstruction on Waymo we report Accuracy, Completeness, and Chamfer Distance against LiDAR ground truth. All experiments use a single NVIDIA RTX 4090 with VGGT~\cite{vggt} as backbone. Full baseline details, per-dataset settings, and implementation specifics are provided in the appendix.

\subsection{Camera Tracking Results}
\label{sec:main_results}

\noindent\textbf{KITTI Odometry.}
\cref{table:kitti_ate} presents results on all 11 KITTI sequences. VGGT-Align achieves the best overall average ATE (19.99) among all methods, ranking first on 7 out of 11 sequences including the longest ones: Seq~00 (4,542 frames, ATE: 7.74), Seq~05 (ATE: 4.97), Seq~08 (ATE: 43.98), Seq~09 (ATE: 23.11), and Seq~10 (ATE: 11.66). Compared to VGGT-Long, our method reduces the overall average by 32\% (29.41$\to$19.99), with up to 54\% reduction on individual sequences (Seq~10: 25.49$\to$11.66). Our Avg$^*$ (7.02) is comparable to VGGT-Long (6.91) while substantially outperforming SwiftVGGT (20.73).

On Seq~01, which involves high-speed highway driving (2.23\,m /frame), all chunk-based methods struggle due to extreme inter-frame displacement. On Seq~02, our ATE (47.04) is higher than the baseline (37.56); this 5\,km sequence is dominated by rotational drift in long loops, which scale anchoring alone cannot fully address.

\begin{figure*}[t]
\centering
\includegraphics[width=0.85\linewidth]{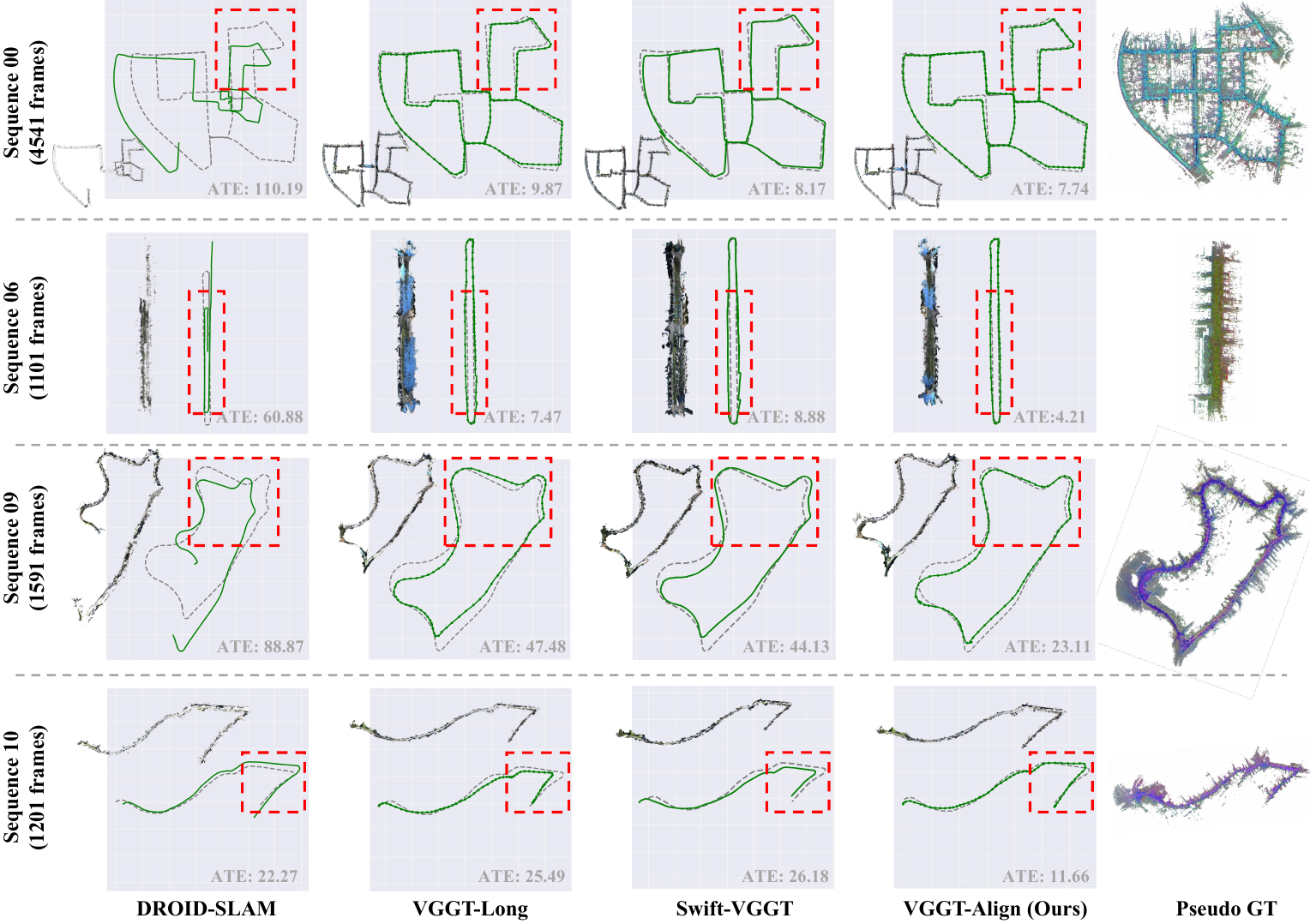} 
\caption{Qualitative comparison of trajectories and dense reconstructions across four KITTI sequences. Columns 1--4: DROID-SLAM, VGGT-Long, SwiftVGGT, and VGGT-Align (Ours); ground truth in gray dashed, predictions in green, red boxes highlight deviations. Column~5: LiDAR pseudo ground-truth point cloud. VGGT-Align achieves the lowest ATE on all four sequences with notable improvements on Seq~06 (62.47$\to$4.21), Seq~09 (72.32$\to$23.11), and Seq~10 (118.70$\to$11.66).}
\label{fig:kitti_traj}
\end{figure*}

\begin{figure}[t]
  \centering
  \includegraphics[width=0.85\linewidth]{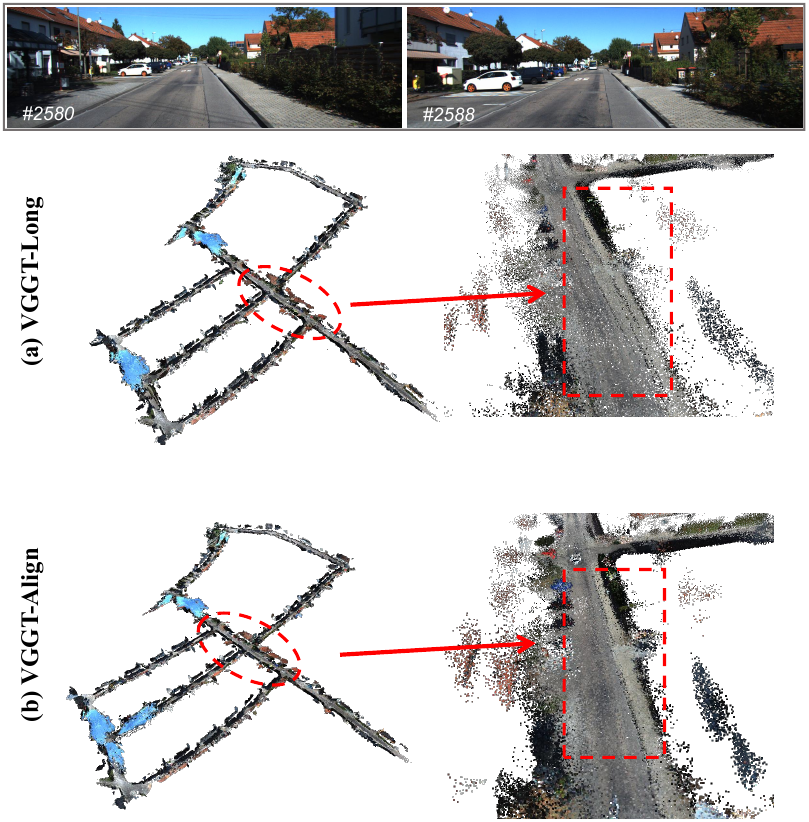}
  \vspace{-10pt} 
  \caption{Dense point cloud comparison on KITTI Seq~05. Top: reference frames (\#2580, \#2588). (a) VGGT-Long: misaligned surfaces, duplicated structures, and scattered outliers at chunk boundaries. (b) VGGT-Align: coherent reconstruction with continuous surfaces and sharp edges from SGIA.}
  \label{fig:pointcloud_cmp}
  \vspace{-5pt} 
\end{figure}

\cref{fig:kitti_traj} provides trajectory and reconstruction visualizations across four sequences. VGGT-Align consistently produces trajectories that most faithfully follow the ground-truth shape. \cref{fig:pointcloud_cmp} further compares dense point clouds on Seq~05: VGGT-Long exhibits misaligned road surfaces and duplicated buildings at chunk boundaries, while VGGT-Align produces geometrically coherent reconstruction with continuous surfaces and sharp edges.

\vspace{4pt}
\noindent\textbf{Waymo Open Dataset.}
As shown in \cref{table:waymo_ate}, VGGT-Align achieves the best average ATE (1.849) among all calibration-free methods, significantly outperforming VGGT-Long (3.085) and SwiftVGGT (2.854) by 40\% and 35\%, respectively. Our method ranks first on 5 out of 9 segments and achieves particularly strong results on geometrically complex segments such as 405841035 (1.020) and 460417311 (1.520). Remarkably, VGGT-Align also surpasses calibration-required DROID-SLAM (4.396), demonstrating that well-constrained chunk-based alignment can effectively rival traditional SLAM systems even without access to camera intrinsics. Trajectory and reconstruction visualizations are provided in the supplementary material.
 
\vspace{4pt}
\noindent\textbf{Virtual KITTI.}
\cref{table:vkitti_ate} evaluates robustness under six appearance conditions on Scene~20, the longest sequence in the dataset (837 frames, 711\,m). VGGT-Align achieves first or second place across all conditions, significantly improving over VGGT-Long. Our calibration-free method even surpasses DROID-SLAM (which requires ground-truth intrinsics) on Morning (3.17 vs.\ 3.73) and Sunset (3.65 vs.\ 4.91), confirming that geometric invariants are appearance-agnostic. Results on additional scenes are provided in the supplementary material.

\begin{table}[t]
  \centering
 \resizebox{\linewidth}{!}{%
\begin{tabular}{l|c|cccccc}
\toprule
 & \textbf{Calib.} & \multicolumn{6}{c}{\textbf{Scene 20}~~~\textitgray{837 frames, 711\,m}} \\
\midrule
\textitgray{Condition}                     & -               & \textitgray{Clone}        & \textitgray{Fog}         & \textitgray{Morning}          & \textitgray{Overcast}        & \textitgray{Rain}       & \textitgray{Sunset}      \\
\midrule
DROID-SLAM \cite{droidslam}                    & \redtext{\textit{Required}}        & \first{3.59}       & \first{5.08}      & \second{3.73}           & \first{3.85}          & \first{3.78}     & \third{4.91}      \\
MASt3R-SLAM \cite{mast3r-slam}                  & \greentext{\textit{No Need}}         & \textitgray{TL}           & \textitgray{TL}          & \textitgray{TL}               & \textitgray{TL}              & \textitgray{TL}         & \textitgray{TL}          \\
CUT3R \cite{cut3r}                        & \greentext{\textit{No Need}}         & 129.50     & 76.96     & 117.95         & 114.51        & 66.70    & 116.53     \\
Fast3R \cite{fast3r}                       & \greentext{\textit{No Need}}         & \textitgray{OOM}          & \textitgray{OOM}         & \textitgray{OOM}              & \textitgray{OOM}             & \textitgray{OOM}        & \textitgray{OOM}         \\
VGGT \cite{vggt}                         & \greentext{\textit{No Need}}         & \textitgray{OOM}          & \textitgray{OOM}         & \textitgray{OOM}              & \textitgray{OOM}             & \textitgray{OOM}        & \textitgray{OOM}         \\
\midrule
VGGT-Long \cite{vggt-long}              & \greentext{\textit{No Need}}         & \third{9.66}       & \third{8.19}      & \third{6.34}           & \third{4.56}          & \third{6.50}     & \second{4.85}      \\
\textbf{VGGT-Align (Ours)}              & \greentext{\textit{No Need}}         & \second{3.70}       & \second{6.20}      & \first{3.17}           & \second{4.27}          & \second{4.68}     & \first{3.65}      \\
\bottomrule
\end{tabular}
  }
  \caption{Camera tracking results (ATE RMSE [m] $\downarrow$) on Virtual KITTI Scene 20 under six appearance conditions. VGGT-Align significantly narrows the gap to calibration-required DROID-SLAM and surpasses it under Morning and Sunset.}
  \label{table:vkitti_ate}
\end{table}

\subsection{3D Reconstruction Quality}
\label{sec:recon}

\cref{table:waymo_point} evaluates dense reconstruction on Waymo. VGGT-Align achieves the best average Accuracy (1.056) and Chamfer Distance (1.541) among all calibration-free methods, confirming that scale-consistent alignment directly translates into higher-quality geometry. Note that LiDAR ground truth has a narrower vertical FoV than RGB cameras, so metrics should be interpreted alongside the qualitative results in \cref{fig:pointcloud_cmp}.

\begin{table*}[t]
  \centering
 \resizebox{0.9\linewidth}{!}{%
\begin{tabular}{l|l|c|c|ccccccccc}
\toprule
\textbf{Methods}  & \textbf{Metric} & \textbf{Calib.}      & \textbf{Avg.} & \textbf{163453191} & \textbf{183829460} & \textbf{315615587} & \textbf{346181117} & \textbf{371159869} & \textbf{405841035} & \textbf{460417311} & \textbf{520018670} & \textbf{610454533} \\
\midrule
\textitgray{Frames / Length (m)}       & \textitgray{-}               & \textitgray{-}                    & \textitgray{198 / 173}       & \textitgray{198 / 160}            & \textitgray{199 / 42}             & \textitgray{199 / 165}            & \textitgray{199 / 351}            & \textitgray{196 / 273}            & \textitgray{199 / 86}             & \textitgray{198 / 266}            & \textitgray{199 / 135}            & \textitgray{198 / 63}             \\
\midrule
\multirow{3}{*}{DROID-SLAM \cite{droidslam}} & Accuracy $\downarrow$           & \multirow{3}{*}{\redtext{\textit{Req.}}}             & \textitgray{1.201}         & \first{0.781}              & 1.136              & 2.247              & 2.393              & \first{1.090}              & 0.539              & \first{0.740}              & \textitgray{TL}                 & \first{0.677}              \\
 & Completeness $\downarrow$           &  & \textitgray{8.540}         & 4.610              & 10.245             & 5.540              & 8.669              & 8.592              & 11.144             & 5.320              & \textitgray{TL}                 & 14.201             \\
 & Chamfer $\downarrow$        &  & \textitgray{4.870}         & 2.696              & 5.691              & 3.893              & 5.531              & 4.841              & 5.842              & 3.030              & \textitgray{TL}                 & 7.439              \\
 \midrule
\multirow{3}{*}{MASt3R-SLAM \cite{mast3r-slam}}          & Accuracy $\downarrow$           & \multirow{3}{*}{\greentext{\textit{No}}}              & 3.772         & 3.189              & 2.988              & 3.787              & 4.689              & 4.436              & 1.166              & 4.637              & 6.417              & 2.637              \\
 & Completeness $\downarrow$          &  & 3.177         & \third{1.715}              & \first{3.284}              & 2.047              & \third{2.981}              & \third{2.679}              & \second{2.895}              & 2.002              & 4.429              & 6.560              \\
 & Chamfer $\downarrow$        &                      & 3.474         & 2.452              & 3.136              & 2.917              & 3.835              & 3.558              & \third{2.031}              & 3.319              & 5.423              & 4.599              \\
 \midrule
\multirow{3}{*}{CUT3R \cite{cut3r}}                & Accuracy $\downarrow$           & \multirow{3}{*}{\greentext{\textit{No}}}              & 3.884         & 3.580              & 1.144              & 2.418              & 3.712              & 3.679              & 4.346              & 2.012     & 12.320             & 1.744              \\
                     & Completeness $\downarrow$          &  & 6.801         & 8.251              & 9.352              & 8.748              & 8.537              & 5.467              & 3.393              & 6.164     & \third{2.302}              & 8.999              \\
                     & Chamfer $\downarrow$        &  & 5.343         & 5.916              & 5.248              & 5.583              & 6.125              & 4.573              & 3.869              & 4.088     & 7.311              & 5.371              \\
\midrule
\multirow{3}{*}{VGGT-Long \cite{vggt-long}}     & Accuracy $\downarrow$           & \multirow{3}{*}{\greentext{\textit{No}}}              & \second{1.182}         & \third{1.002}     & \first{0.395}     & \second{0.925}              & \third{1.668}              & 2.580              & 0.679              & \third{0.784}              & \second{1.358}              & \third{1.246}              \\
 & Completeness $\downarrow$          &  & \third{2.860}         & 2.762     & \second{3.417}     & \third{1.738}              & 3.261              & 2.791              & \third{3.216}              & \third{1.840}              & 4.694              & \first{2.022}              \\
 & Chamfer $\downarrow$        &  & \third{2.021}         & \third{1.882}     & \first{1.906}     & \third{1.331}              & \third{2.465}              & \third{2.685}              & \second{1.948}              & \third{1.312}              & \third{3.026}              & \first{1.634}             \\
\midrule
\multirow{3}{*}{SwiftVGGT \cite{swiftvggt}}     & Accuracy $\downarrow$           & \multirow{3}{*}{\greentext{\textit{No}}}              & \third{1.339}         & 1.175     & 1.364     & \third{1.102}              & \second{1.341}              & 2.622              & \first{0.526}              & 1.256              & \third{1.500}              & \second{1.165}              \\
 & Completeness $\downarrow$          &  & \first{1.985}         & \first{1.049}     & 4.331     & \first{1.317}              & \first{1.353}              & \first{1.716}              & 3.683              & \first{1.236}              & \first{0.963}              & \third{2.221}              \\
 & Chamfer $\downarrow$        &  & \second{1.662}         & \first{1.112}     & \third{2.848}     & \second{1.210}              & \second{1.347}              & \second{2.169}              & 2.104              & \second{1.246}              & \first{1.231}              & \second{1.693}             \\
\midrule
\multirow{3}{*}{\textbf{VGGT-Align (Ours)}}     & Accuracy $\downarrow$           & \multirow{3}{*}{\greentext{\textit{No}}}              & \first{1.056}         & \second{0.964}     & \second{0.489}     & \first{0.777}              & \first{0.835}              & \second{1.713}              & \second{0.537}              & \second{0.774}              & \first{1.191}              & 2.228              \\
 & Completeness $\downarrow$          &  & \second{2.026}         & \second{1.628}     & \third{3.551}     & \second{1.435}              & \second{1.386}              & \second{2.475}              & \first{2.262}              & \second{1.569}              & \second{1.720}              & \second{2.212}              \\
 & Chamfer $\downarrow$        &  & \first{1.541}         & \second{1.296}     & \second{2.020}     & \first{1.106}              & \first{1.110}              & \first{2.094}              & \first{1.400}              & \first{1.171}              & \second{1.455}              & \third{2.220}             \\
\bottomrule
\end{tabular}
  }
  \caption{Dense reconstruction results on the Waymo Open Dataset. Metrics are Accuracy, Completeness, and Chamfer Distance (all in meters $\downarrow$) against LiDAR ground truth. Note that LiDAR has a narrower vertical FoV than RGB cameras, so metrics should be interpreted alongside qualitative results. Color: \first{1st}, \second{2nd}, \third{3rd}.}
  \label{table:waymo_point}
\end{table*}

\subsection{Ablation Study}
\label{sec:ablation}

We ablate each component on Waymo (\cref{table:ablation}). Starting from the VGGT-Long baseline (Avg: 2.154), the ground plane prior alone (GP) does not uniformly help (2.442) due to suboptimal fixed blending. Adaptive blending (AB) reduces this to 2.173. Adding road width prior (RW) brings significant improvement (1.856, $-$14\%), confirming that multi-source invariant fusion is more robust than either source alone. TTA provides a further modest gain, yielding the full model at 1.849. Results on additional scenes are provided in the supplementary material.


\begin{table}[t]
  \centering
 \resizebox{\linewidth}{!}{%
\begin{tabular}{cccc|c|ccccccccc}
\toprule
\textbf{GP} & \textbf{RW} & \textbf{AB} & \textbf{TTA} & \textbf{Avg.} & \textbf{163..} & \textbf{183..} & \textbf{315..} & \textbf{346..} & \textbf{371..} & \textbf{405..} & \textbf{460..} & \textbf{520..} & \textbf{610..} \\
\midrule
\ding{55} & \ding{55} & \ding{55} & \ding{55} & \third{2.154} & \third{1.780} & \third{2.570} & 0.760 & 3.730 & \third{3.225} & 1.390 & \third{1.690} & 3.477 & \first{0.355} \\
\ding{51} & \ding{55} & \ding{55} & \ding{55} & 2.442 & 3.930 & 2.670 & \third{0.588} & \first{2.790} & 3.560 & \third{1.120} & 2.460 & \first{2.060} & 2.800 \\
\ding{51} & \ding{55} & \ding{51} & \ding{55} & 2.173 & 2.540 & 2.600 & 0.616 & 3.360 & 3.320 & 1.190 & 1.830 & \first{2.060} & 2.040 \\
\ding{51} & \ding{51} & \ding{51} & \ding{55} & \second{1.856} & \second{1.290} & \second{2.550} & \first{0.495} & \third{3.280} & \second{2.960} & \first{0.780} & \second{1.530} & 2.190 & \second{1.629} \\
\ding{51} & \ding{51} & \ding{51} & \ding{51} & \first{1.849} & \first{1.160} & \first{2.540} & \second{0.537} & \second{3.140} & \first{2.850} & \second{1.020} & \first{1.520} & \third{2.170} & \third{1.708} \\
\bottomrule
\end{tabular}
  }
  \caption{Ablation on Waymo (ATE RMSE [m] $\downarrow$). Row~1: VGGT-Long baseline. \textbf{GP}: Ground plane Prior. \textbf{RW}: Road Width prior. \textbf{AB}: Adaptive Blending. \textbf{TTA}: Test-Time Adaptation. Row~5: full VGGT-Align. Color: \first{1st}, \second{2nd}, \third{3rd}.}
  \label{table:ablation}
\end{table}

\subsection{Runtime Analysis}
\label{sec:runtime}

We compare runtime against VGGT-Long on KITTI Seq~00 (4,542 frames, 60 chunks, RTX 4090). As shown in \cref{table:runtime}, the VGGT forward pass dominates total runtime and is identical across variants; minor fluctuations ($\pm$0.03\,min) fall within run-to-run variance. Notably, IRLS alignment is \emph{faster} with our modules (1.05\,min $\to$ 0.88\,min), as SGIA provides tighter scale initialization and TTA improves point cloud quality, both yielding cleaner inlier sets and faster convergence. SGIA adds $\sim$0.3\,s/chunk (RANSAC + SVD), totaling $\sim$18\,s ($\sim$5\%); TTA adds $\sim$1.2\,s/chunk (3 steps, 3 frames at 336px), totaling $\sim$72\,s ($\sim$3\%). These costs are largely offset by the IRLS speedup, resulting in net overhead below 3\%, a modest cost well justified by the 32\% ATE reduction.


\begin{table}[t]
  \centering
  \resizebox{\linewidth}{!}{%
  \begin{tabular}{l|cccc|c}
    \toprule
    \textbf{Method} & \textbf{VGGT} & \textbf{Align} & \textbf{SGIA} & \textbf{TTA} & \textbf{Total} \\
    & \textbf{fwd} & \textbf{(IRLS)} & & & \\
    \midrule
    VGGT-Long & 0.68\,min & 1.05\,min & --- & --- & $\sim$2.56\,min \\
    \textbf{Ours} (SGIA) & 0.63\,min & 0.89\,min & 0.14\,min & --- & $\sim$2.31\,min \\
    \textbf{Ours} (full) & 0.62\,min & 0.88\,min & 0.13\,min & 0.07\,min & $\sim$2.32\,min \\
    \bottomrule
  \end{tabular}
  }
  \caption{Runtime breakdown on the Waymo Open Dataset. See Appendix for detailed analysis.}
  \label{table:runtime}
  \vspace{-10pt} 
\end{table}

\section{Conclusion}
\label{sec:conclusion}

We present VGGT-Align, a scale-consistency enhancement framework for chunk-based long-sequence 3D reconstruction. Our core contribution, Scene Geometric Invariant Anchoring (SGIA), exploits cross-chunk consistency of inherent scene geometric quantities to establish scale constraints independent of overlap-based registration, degenerating 7-DoF Sim(3) alignment into 6-DoF rigid-body transformation and severing multiplicative scale drift at its source. Complemented by a lightweight test-time adaptation strategy, VGGT-Align achieves state-of-the-art results on KITTI, Waymo, and Virtual KITTI with less than 8\% runtime overhead. Both modules are plug-and-play and require no offline retraining.



\begin{acks}
This work was supported by the National Natural Science Foundation of China under Grants 62571437 and 62471394.
\end{acks}

\end{document}